\documentclass[lettersize,journal]{IEEEtran} 
\usepackage[utf8]{inputenc}
\usepackage{tikz} 
\usepackage{mathtools}          
\usepackage{acro}               
\usepackage{comment}
\usepackage[ruled,vlined]{algorithm2e}
\usepackage{amsmath}
\usepackage{amssymb}
\usepackage{xcolor}
\usepackage[export]{adjustbox}
\usepackage[colorlinks = true,
            linkcolor = blue,
            urlcolor  = blue,
            citecolor = blue,
            anchorcolor = blue]{hyperref}
\usepackage[font=small]{caption}
\usepackage[caption=false]{subfig}
\usepackage{tabularx,booktabs}
\usepackage{multirow}

\newcolumntype{C}[1]{>{\centering\let\newline\\\arraybackslash\hspace{0pt}}m{#1}}

\begin{document}

\title{Robotic Gas Source Localization with Probabilistic Mapping and Online Dispersion Simulation}
\author{Pepe Ojeda, Javier Monroy and Javier Gonzalez-Jimenez%
\thanks{This work was accepted for publication in IEEE Transactions on Robotics.

\copyright 2024 IEEE.  Personal use of this material is permitted. Permission from IEEE must be obtained for all other uses, in any current or future media, including reprinting/republishing this material for advertising or promotional purposes, creating new collective works, for resale or redistribution to servers or lists, or reuse of any copyrighted component of this work in other works.}}

\maketitle

\begin{abstract}
    Gas source localization (\textit{GSL}) with an autonomous robot is a problem with many prospective applications, from finding pipe leaks to emergency-response scenarios. 
    In this work, we present a new method to perform GSL in realistic indoor environments, featuring obstacles and turbulent flow. Given the highly complex relationship between the source position and the measurements available to the robot (the single-point gas concentration, and the wind vector) we propose an observation model that derives from contrasting the online, real-time simulation of the gas dispersion from any candidate source localization against a gas concentration map built from sensor readings. To account for a convenient and grounded integration of both into a probabilistic estimation framework, we introduce the concept of probabilistic \textit{gas-hit} maps, which provide a higher level of abstraction to model the time-dependent nature of gas dispersion.  
    Results from both simulated and real experiments show the capabilities of our current proposal to deal with source localization in complex indoor environments.
\end{abstract}

\section{Introduction}

Mobile Robotic Olfaction (\textit{MRO)} is a research field that focuses on autonomous robots with the capability of sensing volatile compounds in the air to carry out olfactory-related tasks. \textit{MRO} is an active research field due to its many potential applications, which include detecting dangerous or illegal substances, locating gas pipe leaks, and assisting in rescue missions in inaccessible places. 

The sensory devices that allow for gas sensing are usually referred to as electronic noses, or \textit{e-noses}~\cite{karakayaElectronicNoseIts2020,fengReviewSmartGas2019}, and are often composed of arrays of multiple types of sensors, including gas transducers that are sensitive to different substances, as well as thermometers, hygrometers, \textit{etc}. Another sensor that is important for gas distribution mapping and source localization is an anemometer, since the airflow through an environment greatly impacts the dispersion of any volatiles. 

Two major specific problems are addressed in MRO: gas distribution mapping (\textit{GDM})~\cite{francisreview2022,visvanathanGDM2020a,GONGORA2020655,BURGUES2020} and gas source localization (\textit{GSL})~\cite{chenOdorSourceLocalization2019}. In this paper, we exploit the connection between both problems to design a source localization method that relies building a map of the gas distribution and comparing it to the predictions of a dispersion model.

Gas source localization is particularly challenging because it is a non-observable estimation problem. The sensory information available to the robotic agent --the gas concentration and the airflow vector at one specific point in the environment-- is only indirectly related to the location of the source. Relating both, sensor measurements and the location of the source, requires an observation model that takes into account the complexities of gas dispersion phenomena.
One possible way to derive such an observation model is to rely on predictive dispersion modeling, which takes the source's parameters and boundary conditions of the environment as input to predict the gas concentration at each point in the environment at a future point in time. It is then possible to numerically derive the observation model by comparing the robot sensor measurements with the predictions of the dispersion model when considering all the potential gas source locations.

So far, this strategy has been adopted in previous works by assuming very simple analytical dispersion models such as the Gaussian Plume, and/or making strong assumptions about the environmental conditions: \textit{e.g.} laminar, constant flow, absence of obstacles, known release rates. 

Eliminating these assumptions not only greatly increases the computational complexity of applicable prediction models, but also further complicates the mathematical relationship between the measurements and the potential source locations. In such a scenario, there are many configurations of source position and airflow that give rise to the same concentration value at a given point in the environment. This makes the design of a source observation model based on a single-point measurement intractable.  

\begin{figure*}
    \centering
    \includegraphics[width=0.85\linewidth]{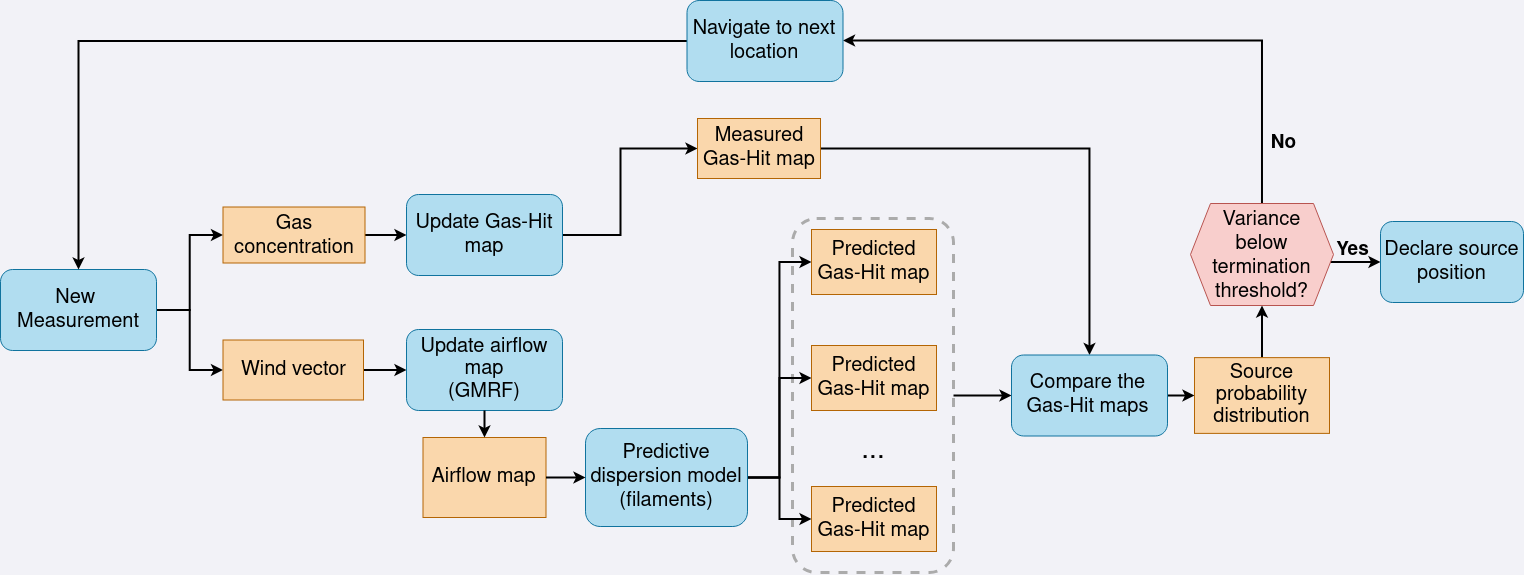}
    \caption{Pipeline of the proposed GSL method. We propose an observation model based on the comparison of the hit-map derived from sensor measurements and those  predicted by a dispersion model.}
    \label{fig:pipeline}
\end{figure*} 

In this work, we tackle this issue with a novel GSL method that brings the following novelties:
\begin{itemize}
    \item A systematic way to estimate the source position from a predictive gas dispersion model. We employ the filament model to allow for real-time computation of the predictions, which makes it possible to dynamically adapt the predictions to the observed environmental conditions (\textit{i.e.} the airflow).

    The results of this predictive model are then used to probabilistically estimate the source location through the method described in Section~\ref{sec:source}. Due to the aforementioned computational complexity of simulating gas dispersion, a crucial part of this contribution is an iterative refinement (coarse-to-fine) method that allows the agent to intelligently allocate its computation time --simulating in detail only the most likely scenarios, and discarding unlikely source positions quickly. 
    
    \item An observation model for the source location that uses a map of the gas dispersion as its observation, rather than single point measurements. 
    Furthermore, this map is abstracted from concentration readings to a more stable and reliable hit-map, where instead of having a continuous concentration value for each cell in the map, we deal with a binary variable representing the presence or absence of gas in that cell. Within the GSL pipeline we treat the resulting hit probability map as a "virtual", more general observation of the gas present in the environment, which ultimately leads to improved performance of the estimation process (explained in detail in Section~\ref{sec:plume-mapping}).
    
    \item An exploration strategy that aims to maximize the information about the source location that is gained with new measurements. We discuss the precedents in this subject, and the main differences between them and our proposal in Section~\ref{sec:movement}.
\end{itemize}

Figure~\ref{fig:pipeline} shows an overview of the structure of the algorithm. Each individual step is explored in detail in Sections~\ref{sec:plume-mapping}-\ref{sec:movement}.


\section{State of the Art}
In this section, we will first give a broad overview of the GSL methods proposed in the past, and then discuss in more detail methods that are directly related to our proposal. For a more general review of the state of the art, see~\cite{francisGasSourceLocalization2022a,lewisComprehensiveReviewPlume2021}.

\subsection{Background on GSL Methods}
Some existing methods frame the problem of GSL in terms of purely reactive navigation, where the sensory input feeds a control loop that steers the movements of the robot~\cite{liAssessmentDifferentPlumetracing2020,macedoComparativeStudyBioInspired2019}. These methods usually rely on either the direction of the airflow (\textit{anemotaxis}) or the direction of the gas concentration gradient (\textit{chemotaxis}) as the main guide for the movement. While this was the most popular way to tackle the problem of source localization for a long time, reactive algorithms have decayed in popularity in recent years, because they cannot cope with the complexity of many real-world situations. For example, in cases where the airflow in the environment is heavily turbulent or time-dependent, no continuous plume or clear concentration gradient may exist, and so the operating principle of these algorithms is invalidated.

More elaborate approaches use the sensory input to estimate a probabilistic belief of the state of the environment, which can include the position of the source itself, the conditions of the airflow, the shape of the gas plume, \textit{etc}. When the estimated variables include more source parameters than just its position, the problem is commonly referred to as \textit{source term estimation} (STE)~\cite{rahbarDistributedSourceTerm2020,hutchinsonReviewSourceTerm2017}. 

These probabilistic solutions offer a more robust way to process the measurements the robot gathers (i.e. wind and gas concentration), since noisy, spurious or unrepresentative measurements are far less likely to throw off the search process. 

\begin{figure*}
    \centering
    \includegraphics[width=0.8\linewidth]{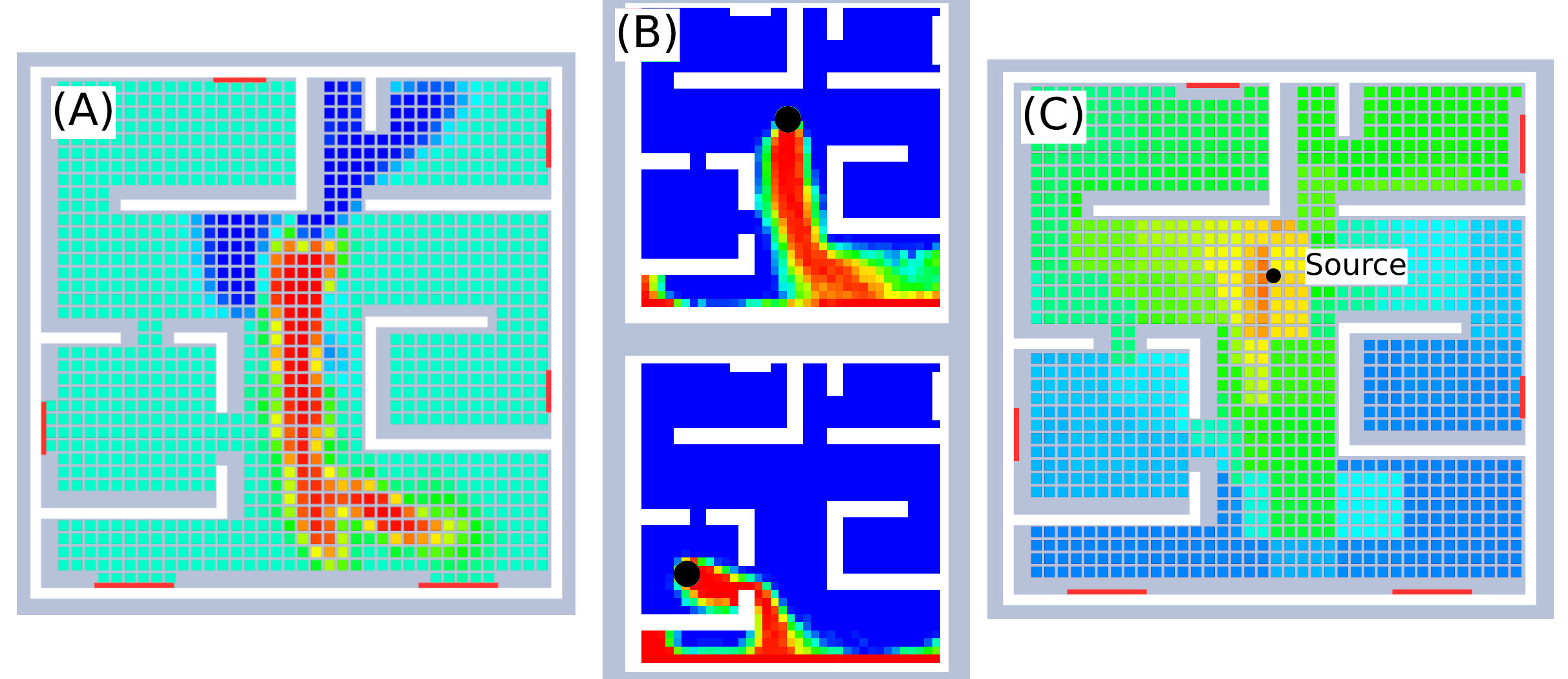}
    \caption{(A) Gas-hit probability map built from e-nose measurements. The color scale goes from blue (lowest) to red (highest). (B) Gas-hit maps predicted by the dispersion model for two candidate source positions (marked in the images with a black dot). (C) Probability distribution of the source location, estimated by comparing A and B.}
    \label{fig:hitP-sourceP}
\end{figure*} 

The main challenge in applying a probabilistic framework is the need for a grounded observation model that relates the state of the environment with the measurements. Both the \textit{e-noses} and the anemometers that robots are often equipped with are single-point sensors and, as was mentioned in the introduction, there is no analytical model that allows to reliably infer the source position from such limited information. Still, multiple options exist, requiring different compromises and offering different advantages. 

For example, if we can assume simplistic environmental conditions (\textit{i.e.} homogeneous airflow, constant gas release rate, absence of obstacles), simple models that define the gas concentration as a function of the sampled position (\textit{e.g.} Gaussian Plume, Isotropic Plume) can be employed~\cite{bourne2019,park2020,magalhaes2020}. With such models, one can infer the location of the source, and even perform STE to obtain extra information about the characteristics of the gas release.

Another possibility is to apply numerical simulations (\textit{i.e.} Computation Fluid Dynamics, \textit{CFD}), which can produce very good estimations of the way gas would disperse under a certain set of environmental conditions. This approach has two main problems. The first one is its computational complexity, as even a single CFD simulation can take hours of computation on a powerful machine. The second one is the fact that CFD models require precise knowledge of the boundary conditions in order to generate an accurate prediction. Some methods~\cite{sanchez-garridoProbabilisticEstimationGas2018,asenovActiveLocalizationGas2019a}\cite{Prabowo2023} have been proposed that get around these limitations by pre-computing a wide range of scenarios and storing the results in a database. During the search, the observations gathered by the robot are contrasted with the simulations in the database to find the most likely one. Despite the superior precision of the models used, this approach presents several important problems. Firstly, it requires the environment to be known, so that all the simulations can be carried out in advance. It also does not scale well with the size or the complexity (number of inlets/outlets, or source positions) of the environment, as the number of simulations that would be required to cover all options dramatically increases.

In a previous work~\cite{Ojeda2021RAL}, the authors presented \textit{GrGSL}, an algorithm which also is designed for source localization in indoors environments with obstacles. This algorithm was based on the propagation of short-range directional estimations by using the geometry of the environment, in a process that treats the occupancy map as a graph and employs a technique loosely based on Dijkstra's algorithm. While this method showed good results in initial testing, the very heuristic nature of its estimation process means that it has difficulty dealing with more complex scenarios (see Section~\ref{sec:experiments}).

Recently, other works have exploited the connection between GDM and GSL to attempt to locate a source by building a map of the gas concentration that is then compared to the predictions of a model that runs in real-time. In~\cite{ercolaniGaSLAMAlgorithmSimultaneous2022}, the authors present an STE method that relies on building a concentration map with the Kernel DM+V/W algorithm~\cite{reggente3DKernelDMAlgorithm2010} and comparing it with the map predicted by a Pseudo-Gaussian plume model --with the limitation that the model does not contemplate the existence of obstacles. In~\cite{jinEfficientGasLeak2023} the authors employ machine-learning with a convolutional neural network trained on CFD results as a surrogate for a gas dispersion model, thus being able to obtain an estimate of the shape of the gas plume for a specific set of source parameters in real time, even in the presence of obstacles.

\subsection{The Filament Dispersion Model}
Our proposal for probabilistic gas source localization relies on a measurement model that stems from the filament-based gas propagation model~\cite{farrellFilaments2002}. It was proposed by Farrell \textit{et al} as a relatively lightweight method for simulating the short time-scale variations in concentration caused by turbulence, as opposed to the time-averaged nature of models like the Gaussian Plume. This model has been used before in the context of robotics, but only for the generation of simulated scenarios~\cite{monroyGADEN3DGas2017}. To the best of our knowledge, this is its first application to online source localization.

There are two main problems that need to be addressed in order to design a source localization method that utilizes the filament model. 

The first one is that despite being much faster to compute than numerical CFD models, running many filament simulations to compare their results with the robot measurements is still a time-consuming task, and naively simulating all potential source locations in even a small environment is not doable in real-time with current hardware. This problem is explored in detail in Section~\ref{sec:coarseToFine}.

The second problem is that filament simulations require knowledge about the airflow in the entire environment, which is not available through sensory measurements, as those are strictly local. Therefore, the application of the filament model to source robotic search requires some method for extrapolating local wind measurements to estimate the global airflow. In an outdoors application, one could make the assumption that the airflow is homogeneous in the search space, even if it changes over time~\cite{asenovActiveLocalizationGas2019a,liOdorSourceLocalization2011}. Indoors, this is not an acceptable assumption, as the presence of walls and obstacles forces the airflow to conform to the geometry of the environment.

\begin{figure*}
    \centering
    \includegraphics[width=0.85\linewidth]{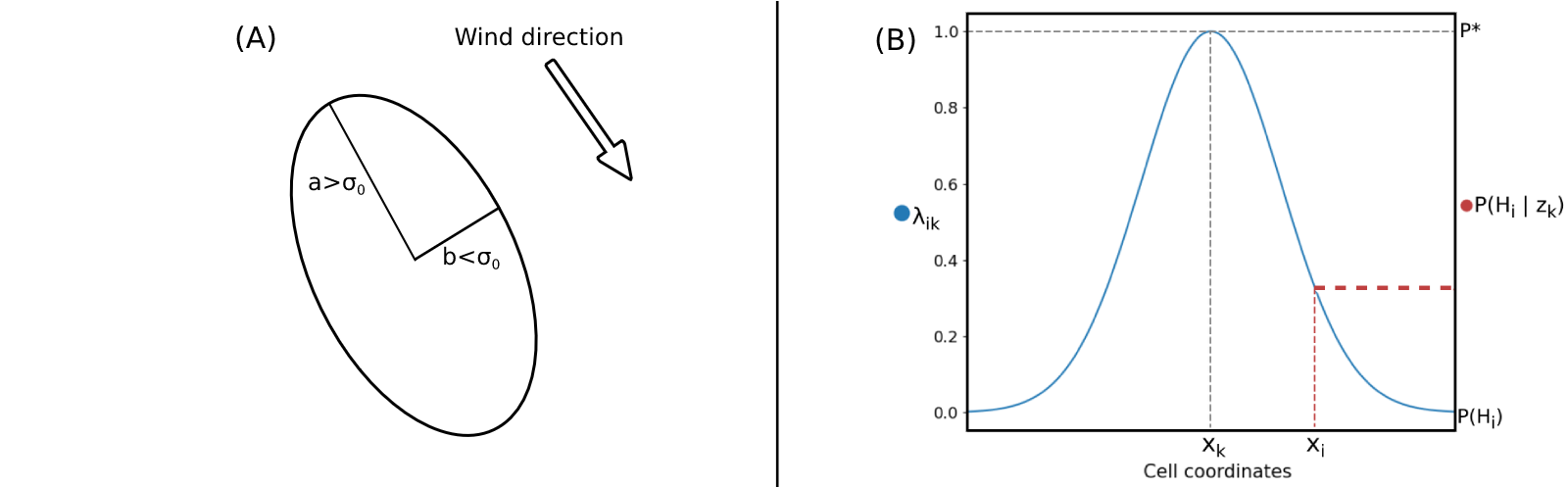}
    \caption{(A) The influence of a measurement ($\lambda_i$) over the probability of $H_i$ is calculated by sampling a 2D gaussian kernel of initial $\sigma = \sigma_0$, centered on the measurement location, that is stretched and rotated by the wind vector. The values of $a$ and $b$ are calculated using the method outlined in~\cite{kerneldm}. (B) One-dimensional simplification of the relation between $\lambda_{ik}$ and $P(H_i | z_k)$, as described by Equation~\ref{eq:conditional_h}.}
    \label{fig:lambda}
\end{figure*}

Since the focus of our current method is on addressing the case of indoors GSL, we will build upon a method proposed by Monroy \textit{et al}~\cite{jmonroy_isoen_2017} to estimate the airflow based on Gaussian Markov Random Fields (\textit{GMRF}). This method estimates the wind vectors over a 2D lattice of cells, imposing constraints on how the airflow can vary from one cell to its neighboring ones and how it must adapt to the shape of the obstacles. By doing this, it is possible to generate a prediction of the direction of the airflow for the entire environment from a few sparse wind measurements. In order to define the constraints necessary to solve for this vector field, this method requires the geometry of the environment to be known, including the positions of possible inlets/outlets (doors, windows, \textit{etc.}). It does not, however, need to know the boundary conditions at those points, or even whether they are actively functioning as inlets/outlets. That is to say that, for example, the map should include the location of all windows, but does not need to specify whether they are currently open.

\section{Probabilistic Plume Mapping}
\label{sec:plume-mapping}
The core idea of the method we present here is to iteratively estimate the source location through the gas distribution mapping (GDM) in the environment. This map is compared to the gas predictions that a filament model generates from a candidate source location, which provides us with a likelihood of the source being in that position (Fig.~\ref{fig:hitP-sourceP}). 

\subsection{Gas Hit Map}

While combining GDM and GSL is not a new idea, it is worth discussing in which ways our current proposal deviates from the conventions and techniques of GDM methods, and the reasoning behind these modifications. The main difference is that, rather than building a grid map of gas concentration values (a continuous random variable), we build a grid map of binary values, where a cell can either contain a measurable amount of gas (a gas \textit{hit}) or not (a \textit{miss}), and the exact concentration value is abstracted away. 
This presence of gas is treated as a random variable, and so we will deal with it in terms of \textit{the probability of a cell containing gas}. Similarly to the \textit{Occupancy Grid Maps} (OGM) employed in mobile robotics\cite{thrun2005probabilistic}, the value at each grid cell represents the probability that the cell contains enough gas to trigger a gas detection event at an arbitrary instant of time. 

From a frequentist perspective, these \textit{hit probabilities} can be interpreted as the proportion of time that a cell contains enough gas to trigger a \textit{hit}. Notice that this concept resembles the idea of plume mapping defined by Farrell \textit{et al.} in~\cite{farrellPlumeMappingHidden2003}, as this probability may be seen as the degree to which each cell belongs to the shape of the time-averaged gas plume: some cells are "stably in the plume" (\textit{i.e.} always contain a concentration of the target gas above a minimum threshold), while others are only partially so, as fluctuations in the shape of the plume can cause them to not contain gas at certain times.

The reason abstract away the concentration values is to better match the degree of precision that is attainable with predictive dispersion models --particularly during an online search, where many of the relevant parameters (boundary conditions, source release rate, \textit{etc}) are uncertain. In this context, the comparison of specific concentration values measured by the sensor with those simulated by the dispersion model is not meaningful, as the simulations cannot be expected to match reality to such a degree of accuracy. Working at a higher level of abstraction attenuates the impact of these limitations --for example, the general shape of a gas plume, as represented by the \textit{gas hit} variable, is largely independent of the release rate of the source, while the concentration values would greatly depend on it.

Formally, a \textit{gas-hit map} is defined as a vector of binary random variables, $H=\{H_i | i \in C \}$, where $C$ is the set of all cells in the environment. We will refer to the set of sensory observations as $Z$, using a subscript to denote the position at which the observation was taken and a superscript to denote the time instant (\textit{e.g.} $z_k^t \in Z$ is the observation taken at time $t$ in cell $k$). These superscripts and subscripts will be omitted for the sake of readability when they are not relevant to the discussion. We will refer to the probability of cell $i$ containing enough gas to trigger a gas hit as $p(H_i)$.

It should be noted that the probabilistic formulation presented in this article is agnostic to the specific definition of \textit{gas hit}. For the purposes of the implementation and experimental validation presented here, a certain observation $z_k^t$ is considered a \textit{hit} or a \textit{miss} simply by comparing the sensor reading to a fixed concentration threshold. However, other approaches which use an adaptive threshold or consider the time-derivative of the concentration reading to define the concept of a \textit{hit} --see, for example~\cite{Li2009}-- could be used.

\subsection{Building the Gas Hit Map}

The estimation of the gas-hit map --$p(H_i|Z) \forall i$-- is carried out recursively by bayesian filtering. Starting from an arbitrary prior $p(H_i)$, each measurement taken in cell $i$ will modify the estimated probability of $H_i$ through the conditional probability $p(H_i | z_i)$. We can define this conditional probability as a piece-wise function that depends on whether the observation is a \textit{hit} or a \textit{miss}:

\begin{equation}
    p( H_k | z_k ) = P^*
    \begin{cases}
        = P_{hit} \text{,   if }z_k =1 \\
        = P_{miss} \text{,   if }z_k =0
    \end{cases}
    \label{eq:p*}
\end{equation}

Because of the complexity of the gas dispersion phenomenon, there is no clear way of setting grounded values for $P_{hit}$ and $P_{miss}$, thus they are left as input parameters to the algorithm, with the only requirement that 

\begin{gather*}
P_{hit} = p(H_k | z_k =1) > p(H_k)\\
\text{and } \\
P_{miss} = p(H_k | z_k =0) < p(H_k)
\end{gather*}

Interested readers can find a similar solution for occupancy mapping in~\cite[p.~33-35]{meraliThesis}. In that work, the author provides a solid discussion on arbitrarily defining conditional probabilities for a binary random variable given sensory observations.

Eq.~\ref{eq:p*} defines the conditional gas-hit probability of a cell only for a measurement taken in the same cell. Building a hit probability map only from this information would require an unfeasible amount of measures. Thus, some kind of inference to neighboring cells is required. For that, we resort to the wind vectors and the known geometry of the environment to compute an estimation of $p(H_i|z_k)$ when $i\neq k$, so a map can be built from sparse measurements.

Intuitively, a hit measured at any specific cell should have no effect on the hit probability of far-away cells, but must change the probability of their surrounding cells according to their distance and position. Concretely, cells that are located upwind or downwind from the measurement location should be more strongly affected, as a noticeable airflow causing advection will create a stronger correlation between the state of cells whose relative position aligns with the wind vector.

Several methods have been proposed in the field of GDM to encode this dependency between cells based on distance and airflow alignment. In this work,  similarly to the Kernel DM+V/W method~\cite{reggenteUsingLocalWind2009}, we apply a dependency model given by a 2D Gaussian centered at the measurement location and aligned with the wind vector. Thus, the influence factor $\lambda_{ik}$ of a measurement taken in cell $k$ over the conditional probability of $H_i$ is given by: 

\begin{equation}
    \lambda_{ik} \propto \mathcal{N}({x_i};{x_k}, \Sigma)
    \label{eq:lambda_no_obs}
\end{equation}

where ${x_i}$ stands for the coordinates of any cell $i$, ${x_k}$ are the coordinates of the cell in which the measurement is taken, and $\Sigma$ is the covariance matrix of the 2D Gaussian, stretched and rotated according to the wind vector as described in~\cite{reggenteUsingLocalWind2009} (see Fig.~\ref{fig:lambda}A). The value of $\lambda_{ik}$ is scaled by $1/N( {0}, \Sigma)$ to  set the maximum value of the influence (when $i=k$) to 1.

We use this influence value $\lambda_{ik}$ to linearly interpolate between two extreme cases for $p(H_i | z_k)$ following the expression:

\begin{equation}
    p(H_i|z_k) = \lambda_{ik} \cdot P^* + (1-\lambda_{ik}) \cdot p(H_i)
    \label{eq:conditional_h}
\end{equation}

\begin{figure}
    \centering
    \includegraphics[width=0.7\linewidth]{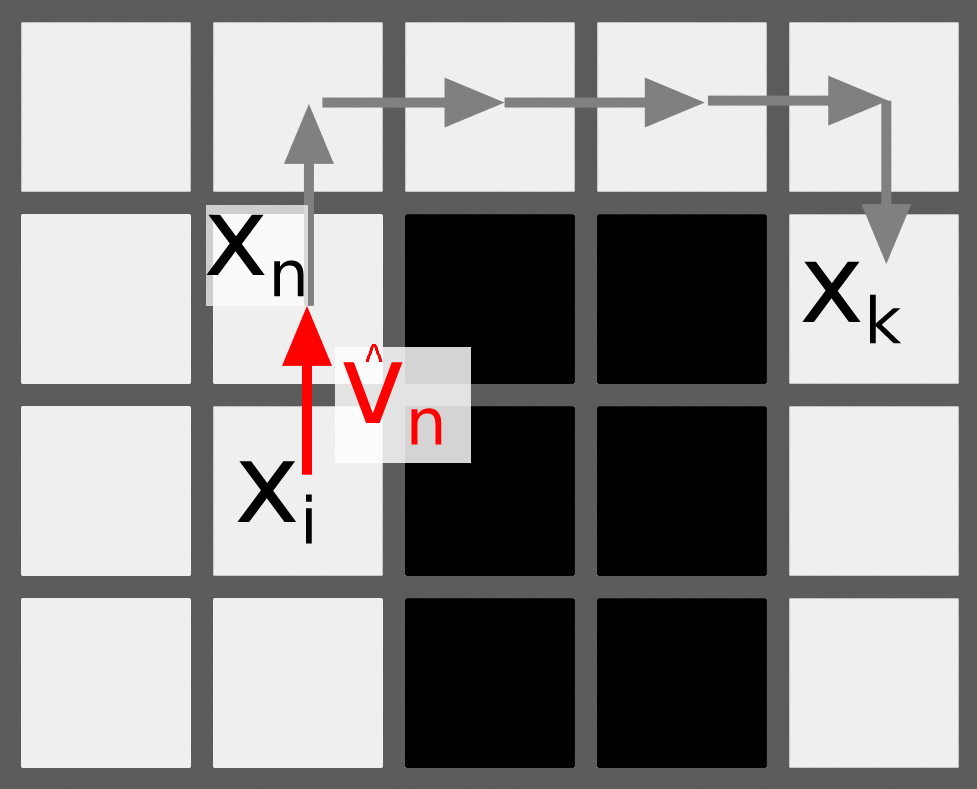}
    \caption{The vector $\hat{v}_n$ used in Eq.~\ref{eq:lamda_obs} is the direction from the considered cell $i$ to the cell $n \in N$ that is part to the shortest path between $i$ and $k$. $\delta_{ik}$ is the total length of said path.}
    \label{fig:lambda_obstacles}
\end{figure}

When $i=k$ (the cell is being directly observed), $\lambda_{ik} = 1$  and we are applying the expression in Eq.~\ref{eq:p*}. When the cell $i$ is very far away from cell $k$,  $\lambda \approx 0$ and $p(H_i | z_k) = p(H_i)$, meaning that $H_i$ and $z_k$ can be considered independent (see Fig.~\ref{fig:lambda}B).

\subsection{Obstacles}
An additional consideration is required for this method to be applicable to real scenarios where obstacles (walls, furniture, etc.) may exist between cells $i$ and $k$. This issue can be addressed by modifying the calculation of $\lambda_{ik}$ in such a way that its value is not based on the relative positions of cell $i$ and $k$, but rather on the direction and length of the shortest free path between them~\cite{visvanathanImprovedMobileRobot2020b}. To efficiently find these paths, we employ the propagation method described in~\cite{Ojeda2021RAL}.

As a brief overview, the method is based on treating the grid of cells as a graph, where each free cell (\textit{i.e.} not occupied by an obstacle) is a vertex, and two vertices are connected if their corresponding cells are both free and neighboring each other. On this graph we apply a search technique based on Dijkstra's algorithm, starting from the set of cells that are neighboring the robot's position ($N$) to find all the cells in the grid which should be assigned to each of the members of $N$ based on traversable distance. The result of this can be seen as a graph partition, where each group of vertices $G_n$ is defined by the vertex $n\in N$ which is closest to all the members of the group. 

Based on this graph partition and the paths calculated in the process of creating it, we can redefine the value of $\lambda_{ik}$ for a cell $i \in G_n$ as:

\begin{gather}
    \lambda_{ik} \propto \mathcal{N}(\hat{v}_n \delta_{ik};\text{   } {x_k}, \Sigma)
    \label{eq:lamda_obs}
\end{gather}

where $\hat{v}_n$ is the vector $({x_n}-{x_k})$ normalized by its length (i.e. unit vector from $x_k$ to $x_n$) and $\delta_{ik}$ is the length of the shortest path connecting $i$ to $k$ (see Fig.~\ref{fig:lambda_obstacles}). Note that this expression becomes the expression in Eq.~\ref{eq:lambda_no_obs} for cells in the immediate vicinity of the robot, but will produce slightly different results for cells that are further away even if no obstacles exist along the path, as the direction of $\hat{v}_n$ might not perfectly line up with $({x_i}-{x_k})$. Implementations of this algorithm may choose to add an extra check for this case and apply Eq.~\ref{eq:lambda_no_obs} whenever a direct line of sight exists, but our testing shows no difference in the effectiveness of the algorithm in either case.

\subsection{Bayesian Update}

We have so far omitted the time superscript on the measurements, and described only the conditional probability of $p(H_i | z_k^t)$, for a single measurement $z_k^t$. However, the actual estimation of the map of hit probabilities is based on combining the information obtained from all the accumulated measurements. This process can be done recursively through Bayesian filtering, where we consider the conditional probability discussed in the previous sections, $p(H_i | z^t)$, as the inverse sensor model.

Since the variable $H_i$ we are interested in when building the map is a binary random variable, we can use the log-odds form of the binary Bayes filter~\cite{thrun2005probabilistic}:

\begin{equation}
    l(H_i|z^{1:t}) = l(H_i | z^{1:t-1}) + l(H_i | z^t) - l(H_i)
\end{equation}

where $l(x)$ denotes the log-odds:
\begin{equation*}
 l(x) = \frac{p(x)}{1-p(x)}   
\end{equation*}
and the actual probability value can be recovered with:
\begin{equation*}
 p(x) = 1-\frac{1}{1+e^{l(x)}}
\end{equation*}

\subsection{Confidence Value}
An important consideration when using the generated hit probability map to estimate the location of the source is that the hit probability values that are estimated at some locations are based on measures in the near vicinity of that cell, while others are only due to extrapolation, or even still equal to the uninformative prior. Trivially, the estimations about the presence of gas that are based on extensive observations should have a stronger effect over the predicted source location probabilities.

Thus, it becomes necessary to quantify the uncertainty about the hit probabilities at each cell, which can be defined as a function of how many measurements have been gathered, and how close to the cell those measurements were taken. In this work, we will use the confidence measure $\alpha$ introduced in~\cite{kerneldm}, which is calculated as follows:

\begin{gather}
    \Omega_i = \sum_t \mathcal{N}(\delta^t; 0, \sigma) \nonumber\\
    \alpha_i = 1-e^{-\Omega_i^2 / \sigma_\Omega^2}
    \label{eq:confidence}
\end{gather}

where $\delta^t$ is the length of the shortest path between $x_i$, (the position of the cell being updated), and $x_z^t$ (the position at which the robot took a measurement at timestep $t$). Both $\sigma$ and $\sigma_\Omega$ are parameters that control how much confidence is gained from each individual measurement. For more information about these parameters, see~\cite{kerneldm}.

\section{Source Position Estimation}
\label{sec:source}
In this section, we will discuss the process of using the map of $p(H_i|Z) \forall i$ that was built from the measurements to generate an estimation of the position of the source.

\subsection{Hit Map Comparison}
We define a random variable $S$ to represent the source location as a discrete variable whose possible values are each of the free cells in the environment. The probability distribution that we want to calculate is thus $p(S|H)$. 

As explained in Section~\ref{sec:plume-mapping}, the core idea of our proposal is that we can estimate the probability of a certain cell being the source location by comparing the plume that would result from having the source there (according to a dispersion model), to the plume we have constructed from measurements. Specifically, when we talk about "comparing the plumes", we mean comparing the \textit{hit probabilities}, where these probabilities can be understood as relative frequencies. We define $f^z_i = p( H_i | Z)$ and $f^{S_k}_i = p(H_i | S_k)$ as these predicted relative frequencies given the measurements and given the predictive model for a source in cell $k$, respectively. The absolute difference between these two values, $\Delta_{ik}$, can then be used as a measure of the similarity between the measured and the predicted states of cell $i$:

\begin{equation}
    \Delta_{ik} = |f^z_i-f^{S_k}_i|
\end{equation}

We can then define the conditional probability of $S$ with the following expression:

\begin{gather}
    p(S_k|H_i) \propto \alpha_i \cdot  (1- \Delta_{ik}) + (1-\alpha_i) \cdot 1
\end{gather}

A value of $f^z_i$ that is based on a very low confidence estimation ($\alpha_i \approx 0$) does not give meaningful information about the source, and so the conditional probability distribution of the source is uniform -- $P(S_k | H_i) \propto 1 \text{   } \forall k$. As $\alpha_i$ approaches 1, the probability distribution of the source favors the locations which predict a value of $f^{S_k}_i$ similar to $f^z_i$.

Assuming conditional independence of each $H_i$ given $S$, the probability distribution of the source is then calculated as: 

\begin{multline}
    p(S_k|H) = \prod_i \frac{ p(S_k|H_i) \cdot p(H_i)}{ p(S_k) \cdot p(H_i | H_{\neg i} ) } = \\
    \prod_i p(S_k|H_i) \cdot \prod_i \frac{p(H_i)}{ p(S_k) \cdot p(H_i | H_{\neg i} ) }
\end{multline}

where $H_{\neg i}$ denotes the set $\{H_j | j \neq i\}$. If we assume the prior probability distribution of the source location to be uniform, the term $\prod_i \frac{p(H_i)}{ p(S_k) \cdot p(H_i | H_{\neg i} )}$ is equal for all $k$, and since we assume there is a single gas source, we can simply omit it from the calculation and normalize the resulting values to obtain a valid probability distribution:

\begin{equation}
    p(S_k|H) \propto \prod_i p(S_k|H_i)
\end{equation}

\begin{figure*}
    \centering
    \includegraphics[width=\linewidth]{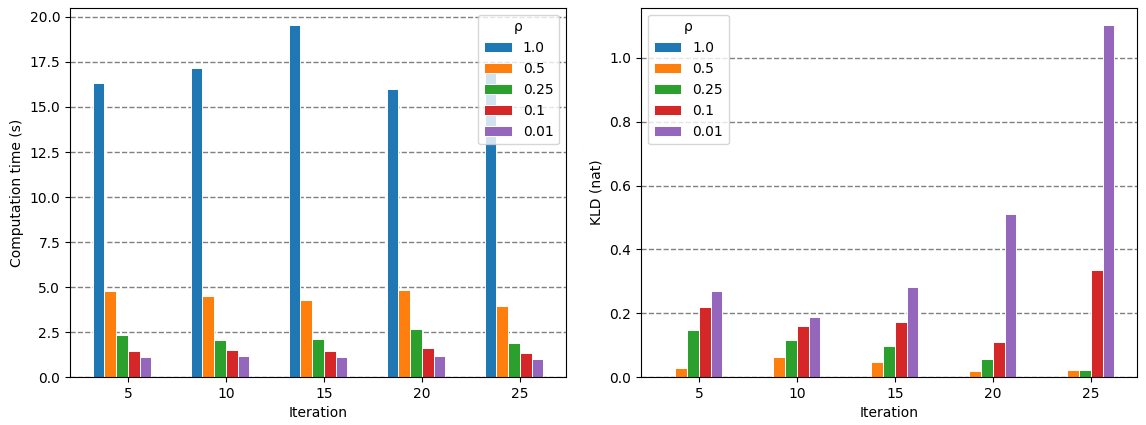}
    \caption{Effect of modifying the fraction of cells that are subdivided on each coarse-to-fine step ($\rho$). (A) Computation time for updating the source probability distribution. (B) Kullback-Leibler divergence of the resulting source location probability distribution with respect to the distribution predicted by considering all the cells.}
    \label{fig:refinement}
\end{figure*}

\subsection{Filament model}
The formulation discussed in the previous section requires a predictive gas dispersion model that, given a source position, produces a map of hit probabilities for the expected gas plume. Because of the types of environmental conditions that we are considering (indoor environments with obstacles), we opt for a simplified version of the filament model.

In the original filament model, as proposed by Farrell \textit{et al}~\cite{farrellFilaments2002}, the dispersion of gas is simulated by tracking the movement of discrete units (the filaments), which in turn represent three-dimensional spatial distributions of concentration --usually modeled as normal distributions centered on the filament's position. The filaments are assumed to move mostly through advection, and the effects of diffusion are accounted for by changing the parameters of the concentration distribution that each filament represents. That is, a filament that was emitted a long time ago represents a normal distribution of higher $\sigma$ than one that was just released from the source, even though both distributions contain the same number of moles of the realized gas.

For the purposes of this work, the filament model has been simplified in two ways. The first and most important one is that the simulation takes place in two dimensions rather than three. This is partially to alleviate the computational complexity of the proposed method, but also to account for the fact that only 2D airflow information is available from the GMRF estimations.

The second way in which our implementation deviates from the filament model is that, since we are not interested in computing concentration values (for the reasons discussed in Section~\ref{sec:plume-mapping}), we are not modeling the filaments as normal distributions of concentration. Instead, we consider that filaments have a discrete radius that grows as a function of the length of time that has passed since their emission. During the simulation, we record the number of time instants that each cell is close enough to the position of at least one filament to be occupied by gas. The relative frequency of this event is the value of $f^{S_k}_i = P(H_i | S_k)$.

\subsection{Iterative Coarse-to-Fine Refinement}
\label{sec:coarseToFine}
Despite the adoption of a simplified model, the filament model still has a significant computational cost, as a complete simulation needs to be carried out for each cell in the environment. In order to alleviate this computational load, some optimizations can be implemented so that a higher proportion of the time is spent on the calculations that will significantly impact the algorithm's predictions about the source location.

Consider, for example, a simulation with the source in cell $i$ whose predicted plume map does not match at all the measurements that have been taken so far. It is trivial that simulations with the source in the immediate neighbors of $i$ will produce similar plumes, which will also be poor matches for the real measurements, and thus those cells will be evaluated as unlikely source locations. On the contrary, areas of the map that are evaluated positively benefit much more from increased resolution, as it may be required to differentiate from several likely source candidates. We propose to utilize this idea by employing a progressive refinement (coarse-to-fine) strategy, where only a few source locations are simulated initially as representatives of coarse regions, and those regions which produce the most promising results are recursively subdivided to obtain more precision in the final estimations. 

\begin{figure}
    \centering
    \includegraphics[width=\linewidth]{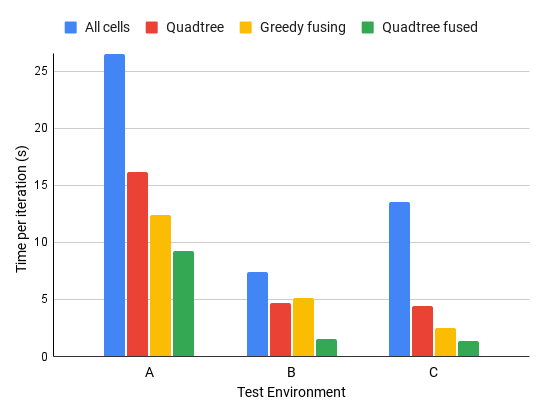}
    \caption{Average time required to update the source position distribution with the discussed optimizations in three different environments, with $\rho=0.5$}
    \label{fig:optimization}
\end{figure}

We refer to the proportion of cells that gets subdivided in each successive step of this process as the refinement fraction ($\rho$). That is, for $\rho = 0.5$, half of the cells (the half with the highest estimated probability of containing the source) are subdivided for finer simulation in the next step. Figure~\ref{fig:refinement} shows how $\rho$ affects the computation time for updating the source's probability distribution and the quality of the resulting estimation as the search process progresses. The quality of the estimation is measured as the Kullback-Leibler Divergence (\textit{KDL}) between the probability distribution of $S$ obtained with that refinement fraction, and the probability distribution of $S$ that results from considering all cells. Note that the KLD for $\rho=1$ is trivially always 0, as its resulting probability distribution is the one being approximated.

It can be observed that for $\rho> 0.25$ the KLD becomes negligible as the number of iterations of the algorithm advances. Conversely, the results obtained with a very small refinement fraction worsen over time. This effect is caused by the variance of the source's distribution decreasing. As the search advances, a small number of nearby cells accumulate most of the probability of containing the source. When this happens, subdividing that small area of the map with a high probability of $S$ is enough to obtain a probability distribution very close to the one being approximated. For very small refinement fractions, the KLD increases because $P(S_k|H)$ can vary more drastically for nearby cells at this stage of the search than when only part of the map has been observed, and thus the very limited amount of subvidisions is not enough to capture the high-frequency variations in the estimated probability of the source location. 

We tested two different methods for generating these regions, both of which produce rectangular cells and allow for a maximum cell size to be specified to make sure the regions are not so coarse as to have the environmental conditions change significantly inside of them. The first method is to generate a quadtree from the occupancy map, which allows for large unoccupied regions to be grouped together into a single leaf of the tree, while allowing obstacles to serve as boundaries that force areas to be considered separately. The second method is to greedily fuse the free cells in the occupancy map with their neighbors, in an arbitrary order, with the only constraint being that the resulting cells must remain rectangular. While both of these methods allowed for a significant speed increase from the naive approach (see Figure~\ref{fig:optimization}), the biggest speedup was obtained by combining both: generating a quadtree, and then fusing the resulting neighboring leaves.

Note that none of these methods are guaranteed to generate an optimal number of subdivisions of the map according to our constraints. A more in-depth study of this problem is left for future work.

\section{Movement Strategy}
\label{sec:movement}
In this section we will address the subject of selecting the next position where the robot will take a measurement, including discussion of existing literature and areas of potential future improvement.

\subsection{Information Value}
\label{sec:movement-information}
One of the most common strategies for planning the movements of the robot is to maximize the information about the source location that is gained with each new measurement. This idea has been extensively explored by previous methods~\cite{AN2022,ZHAO2020,hutchinson2018}, in what is usually referred to as "information-theoretic" movement strategies. One of the most notable examples of this idea is Infotaxis~\cite{vergassola2007}, which popularized the use of the expected change in the entropy of the source distribution as a measure for the information gain. Later works~\cite{hutchinson2018,Ojeda2021RAL} have proposed using the Kullback-Leibler Divergence instead, since aiming only to decrease entropy can cause the robot to refuse exploration and only focus on confirming its current belief. Some recent works~\cite{AN2022,rhodes2023} have proposed combining the Infotaxis paradigm with rapidly-exploring random trees (RRT) as a way to handle the presence of obstacles for selecting the new sampling location.

All of these approaches have a common problem: they require estimating what the next measurement will be if the robot moves to a given position. With this hypothetical next measurement, they simulate an update on the source position belief, and compare the result to the current one. This requires a method for reliably estimating what the next measurement will be, which is not trivial, and also requires performing an iteration of the source estimation procedure for each considered movement, which can lead to a prohibitive computational cost if the number of possible movements is large. Since our algorithm has a slow update process, requiring many filament simulations, this approach is not feasible.

\begin{figure}
    \centering
    \begin{subfloat}
        \centering
        \includegraphics[width=\linewidth]{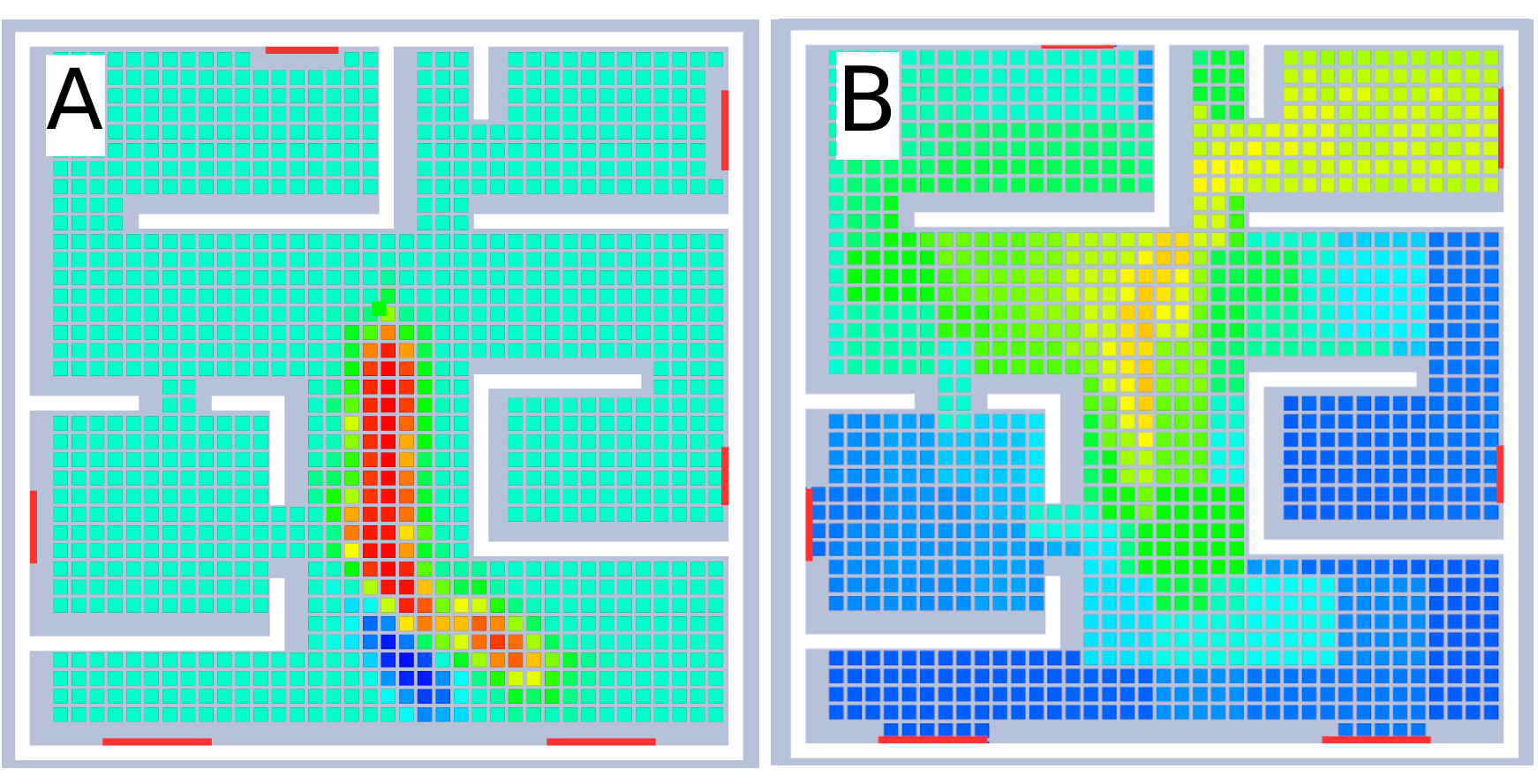}
    \end{subfloat}
    
    \begin{subfloat}
        \centering
        \includegraphics[width=0.27\linewidth,valign=c]{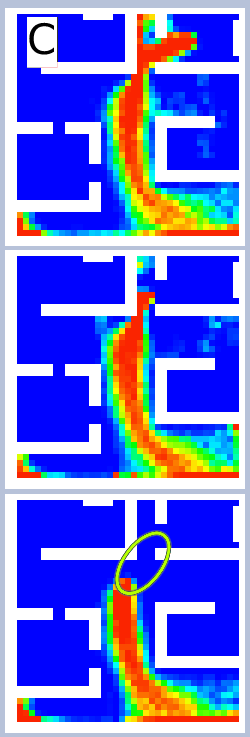}
    \end{subfloat}    
    \begin{subfloat}
        \centering
        \includegraphics[width=0.5\linewidth,valign=c]{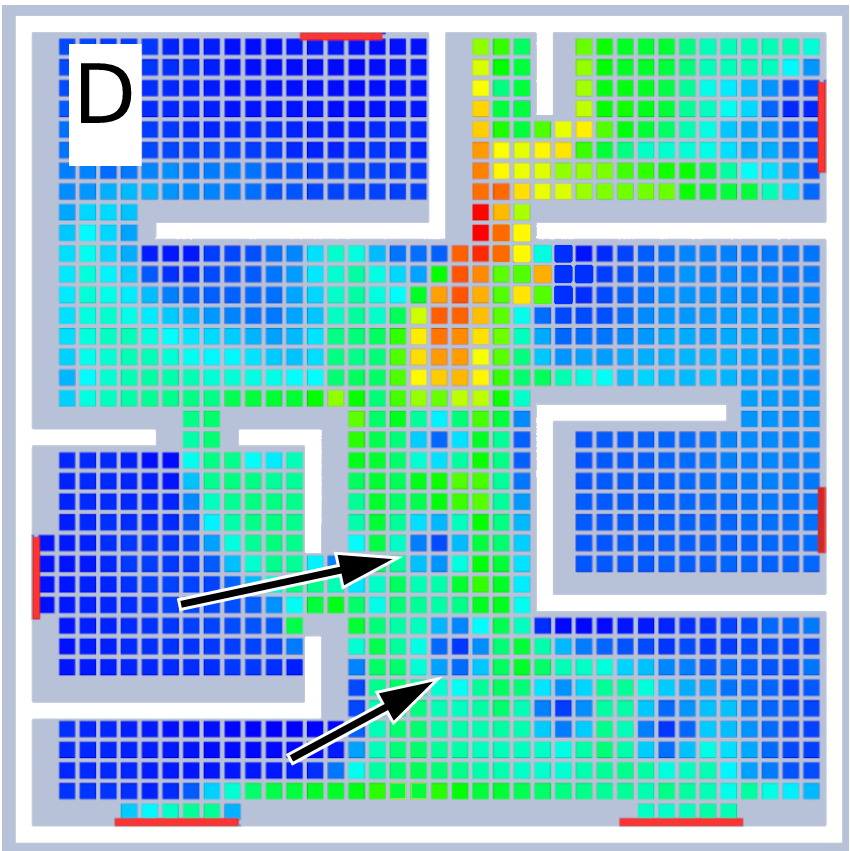}
    \end{subfloat}
    
    \caption{The movement strategy tries to maximize the information about the source location by visiting the locations where the expected plumes vary the most. (A) Current hit probability map. (B) Current source probability distribution. (C) Examples of the maps generated by the dispersion model for the most likely source locations. The area marked with an ellipse changes has the highes variation so it is chosen as the next measurement location. (D) Information value estimated for each possible next sampling position ($\psi_i$).}
    \label{fig:movement}
\end{figure}

Instead, we propose using the already-computed filament simulations to identify the most interesting areas. When trying to discriminate between two possible source positions, the areas of most interest for taking measurements are those in which their respective predicted plumes are most different (see Figure~\ref{fig:movement}). We generalize this to all possible source positions and use their estimated probabilities of containing the source as a weight for how much their prediction influences the interest of a measurement location.  This is represented by the variance of $p(H_i|S)$. The information value of a particular measuring location, $\psi$, is therefore calculated as follows:

\begin{gather}
    \psi_i = (1-\alpha_i) \cdot \sum_k p(S_k | H) \cdot (f_i^{S_k} - \mu_i)^2 \\
    \text{with  } \mu_i = \sum_k f_i^{S_k} \cdot p(S_k | H) \nonumber
\end{gather}

The term $\mu_i$ is the expected value of the predicted relative frequency of hits in cell $i$ ($f_i^{S}$), calculated as the average of $f_i^{S_k}$ weighted by the estimated probability of each source location $k$. This expected value is then used to compute the variance of $f_i^{S}$, thus quantifying how much the presence of gas in cell $i$ depends on where the source is located.

The resulting variance is finally multiplied by the term $(1-\alpha_i)$, where $\alpha_i$ is the confidence value introduced in Eq.~6. This serves to make areas already visited by the robot less interesting, which is desirable because it avoids having the robot constantly re-observe areas of low uncertainty, where no more information about the source can be obtained.

It should be noted that the value of $\psi_i$ represents only the expected amount of information to be gained, and does not consider the cost of selecting cell $i$ as the next location --where this cost could be defined simply in terms of the time required for the robot to navigate to the selected cell. Therefore, even if the value of $\psi$ is assumed to be an accurate estimation of the information that will be gained, there is no guarantee that always moving to maximize $\psi$ will lead to the fastest possible convergence of the algorithm, even though it should minimize the number of required iterations.

The subject of balancing information gain and navigational cost has been extensively explored in the literature, for example for the problem of generating occupation maps~\cite{colaresNextFrontierCombining2016,fangAutonomousRoboticExploration2019,gonzalez-banos2002}. Thus any existing method could be applied to the case of this algorithm. A detailed analysis of the results obtained with these methods, as well as the effect of assigning specific values to their input parameters, is beyond the scope of this paper, and is left for future work. For the purposes of the experimental results presented in the next section, the strategy employed is a greedy maximization of the expected information value.

\subsection{Exploration Phase}

A problem with using the previously defined information gain metric, $\psi$, is that it relies on being able to apply the dispersion model to predict the shape of the gas plumes. However, since no information about the boundary conditions of the environment is provided to the algorithm, the estimation of the airflow during the first few iterations is not reliable, and so is the filament model.

We have tackled this limitation by introducing an initial exploration step of an arbitrary number of iterations, defined as an input parameter of the algorithm-- during which the robot always moves to maximize the total observed area. To do this it uses $\alpha_i$, the confidence value, of both the cell being considered as the next location and all other cells in its neighborhood to compute which movement has the best exploration value. The expression applied to calculate this exploration value is as follows:

\begin{equation}
    \psi_i^{exploration} = \sum_{n \in N_i} (1-\alpha_n) \cdot e^{-\delta_{in}}
\end{equation}

where $N_i$ is the set of cells that are near cell $i$ --for an arbitrarily defined distance limit--, and $\delta_{in}$ is the distance of the shortest path between cells $i$ and $n$. In practice, the distance limit is just a computational optimization, as the influence of other cells on the evaluation quickly approaches 0 as $\delta_{in}$ increases.
        
In the process of trying to cover the largest possible area, the robot will gather sparse measurements of both gas and airflow that span a significant portion of the map. Thus, when the exploration phase ends, both a rough outline of the gas map, and a tentative estimation of the airflow are present, making the source estimation process possible.

It should be noted that this exploration process is not optimal, as it does not actively consider the airflow estimation, but simply how much the observation will expand the amount of observed area. Designing a more optimized strategy is challenging and beyond the scope of this paper, as it would require predicting how future measurements would change the predicted airflow map after the GMRF is updated.



\section{Experimental Validation}

\subsection{Accuracy of Hit Maps}

As discussed in Section~\ref{sec:source}, the probability distribution for the source location is derived from the similarity between the measured and simulated gas hit maps. However, there is an open question as to whether those maps accurately reflect the state of the gas dispersion in the environment. In this section, we will look at a few example environments, where these reconstructed hit maps are compared to the ground-truth hit frequency extracted directly from a full 3D, CFD-based simulation.
                
\begin{figure}
    \centering
    \includegraphics[width=\linewidth]{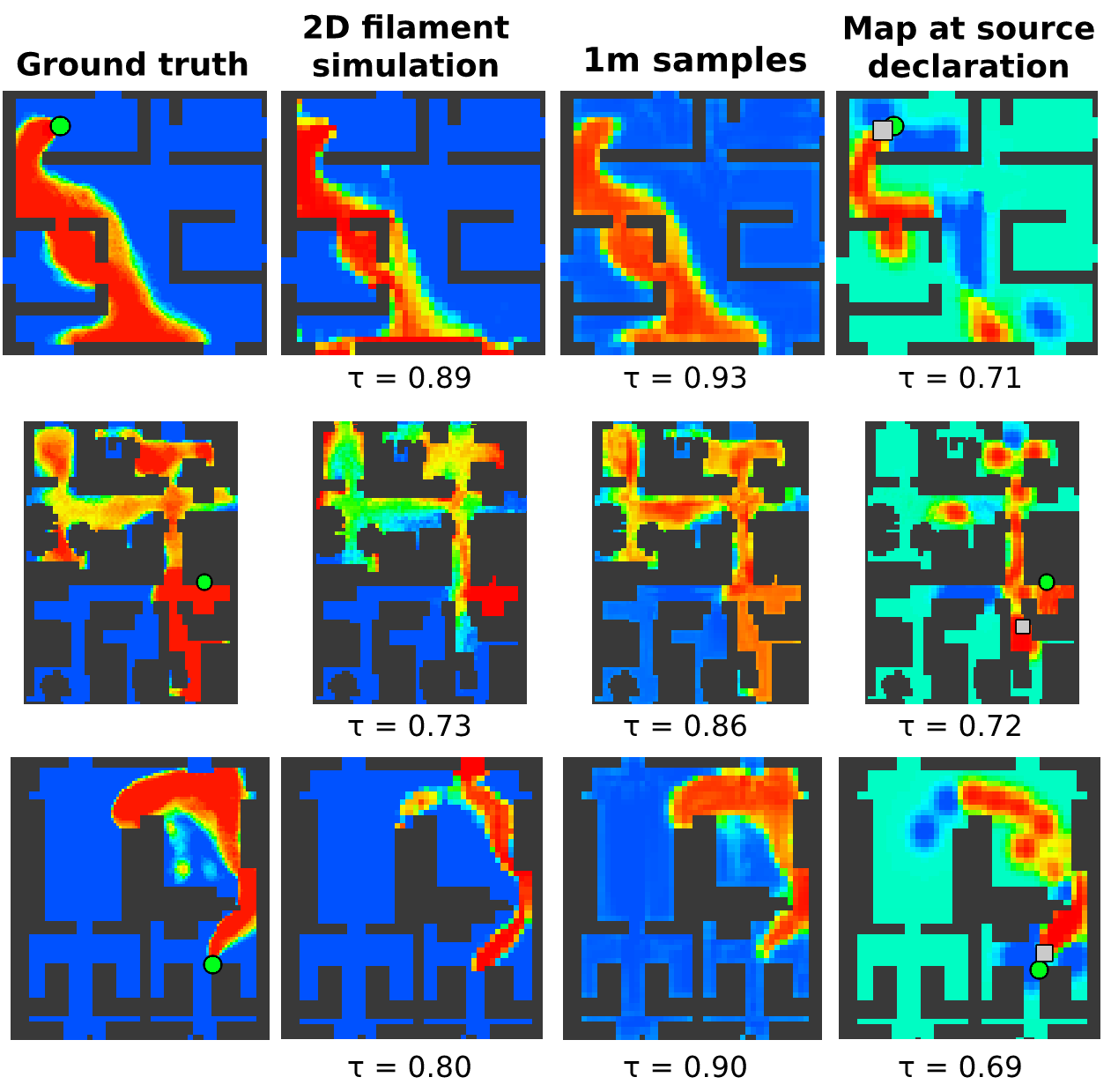}
    \caption{Comparison of the hit maps involved in the algorithm. Column 1 shows the ground truth value, extracted from Gaden, which uses a CFD simulation and the full 3D filament model. Column 2 shows the predicted hit map using the online simplified 2D filament model and reconstructed airflow. Column 3 shows the map produced from sensory measurements taken at regular 1 meter intervals. Column 4 shows the map generated during the GSL process. The real source location is marked in columns 1 and 4 with a circle, and the final declared source position appears in column 4 with a square.}
    \label{fig:hitmaps}
\end{figure}    
Figure~\ref{fig:hitmaps} presents three different environments for comparison purposes, corresponding to experiments A2, B2, and E2. The leftmost column showcases the reference map extracted from the CFD simulation, while each subsequent column displays a map generated by our method, intended for comparison with the reference map. Underneath each map, the similarity measure $\tau$ is shown, calculated as $\tau = 1- \frac{1}{N} \cdot \sum_i^N |f_i - GT_i|$, where $f_i$ represents the estimated hit frequency in a given cell $i$, $GT_i$ denotes the corresponding value in the ground-truth map, and $N$ is the total number of cells.

The second column shows the hit maps generated by the simplified 2D filament model, with airflow reconstructed using the GMRF technique~\cite{jmonroy_isoen_2017}. Notably, in experiment A2, the simulated hit map closely matches the ground truth, while experiments B2 and E2 display more significant differences. These disparities mainly arise from the more pronounced three-dimensionality of airflow in these scenarios, which 2D filament model cannot capture. Experiment B2, in particular, exhibits the largest deviation, where the model predicts an absence of gas in the bottom-right corner, contrary to reality. This sheds light on the results presented in Section VI, where experiment B2 is highlighted as having the highest error in the source position declared by our method.

Moving to the third column, we see the hit maps as reconstructed from distributed sensory measurements taken at 1m-spaced grid. This validates that, with a sufficient number of measurements, our method accurately reconstructs the gas plume's shape. It can be observed that in all cases depicted the results closely align with the ground-truth map.

Finally, the last column illustrates the gas hit maps reconstructed from measurements during a search process for the source, specifically at the moment of source declaration. It's worth noting that in all cases, a significant portion of the map remains unobserved, with the hit probability remaining equal to the prior. This emphasizes our method's capability to declare the source position without requiring a perfect, comprehensive gas map, allowing for faster and sparser exploration.

\begin{figure*}[t]
    \centering
    \includegraphics[width=0.8\linewidth]{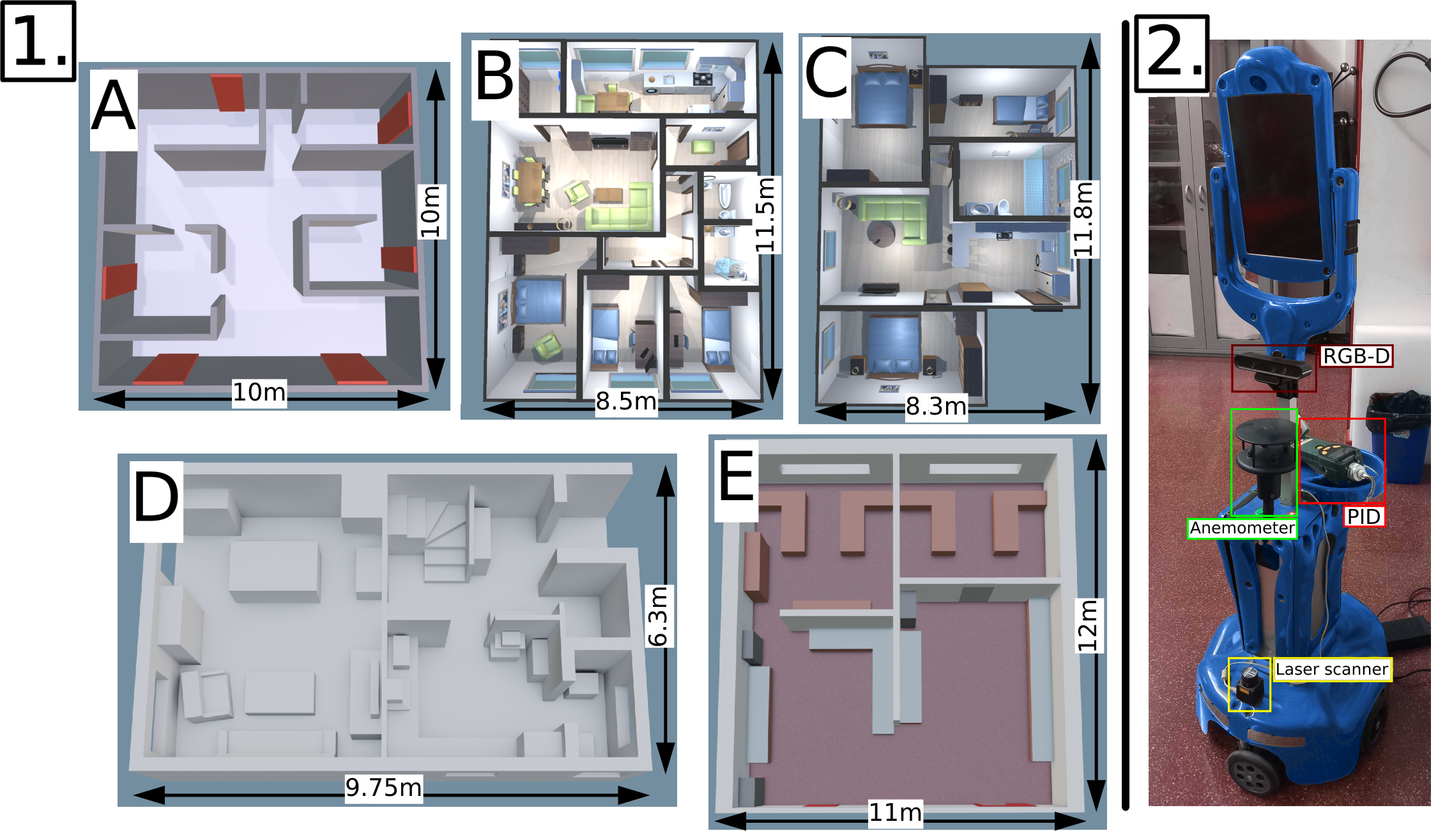}
    \caption{(1) 3D models of the scenarios in which the experiments take place. Details of the environment configuration for each experiment can be found in the online repository. (2) The robot and sensory equipment utilized for the real-world experiments.}
    \label{fig:sim_experiments}
\end{figure*}

\begin{figure*}
\centering
\begin{minipage}{\textwidth}
    \centering
    \includegraphics[height=0.45\textheight]{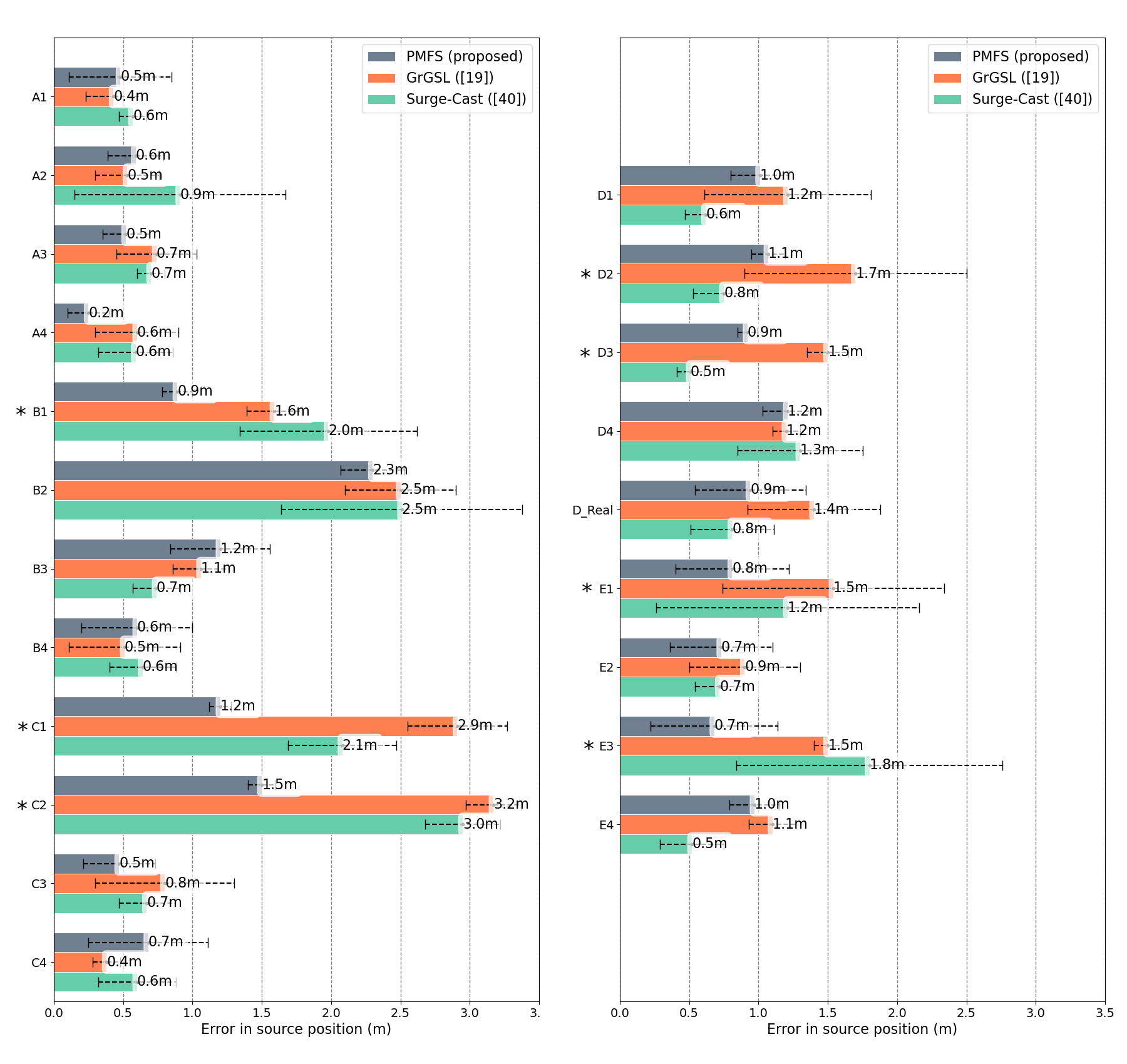}
    \caption{Average error in meters between the final estimated source location and the ground truth in each of the experiments. For Surge-Cast --lacking source declaration--, the reported error is the smallest distance achieved between the robot and the source position.}
    \label{fig:results_error}
\end{minipage}
\begin{minipage}{\textwidth}
    \centering
    \includegraphics[height=0.45\textheight]{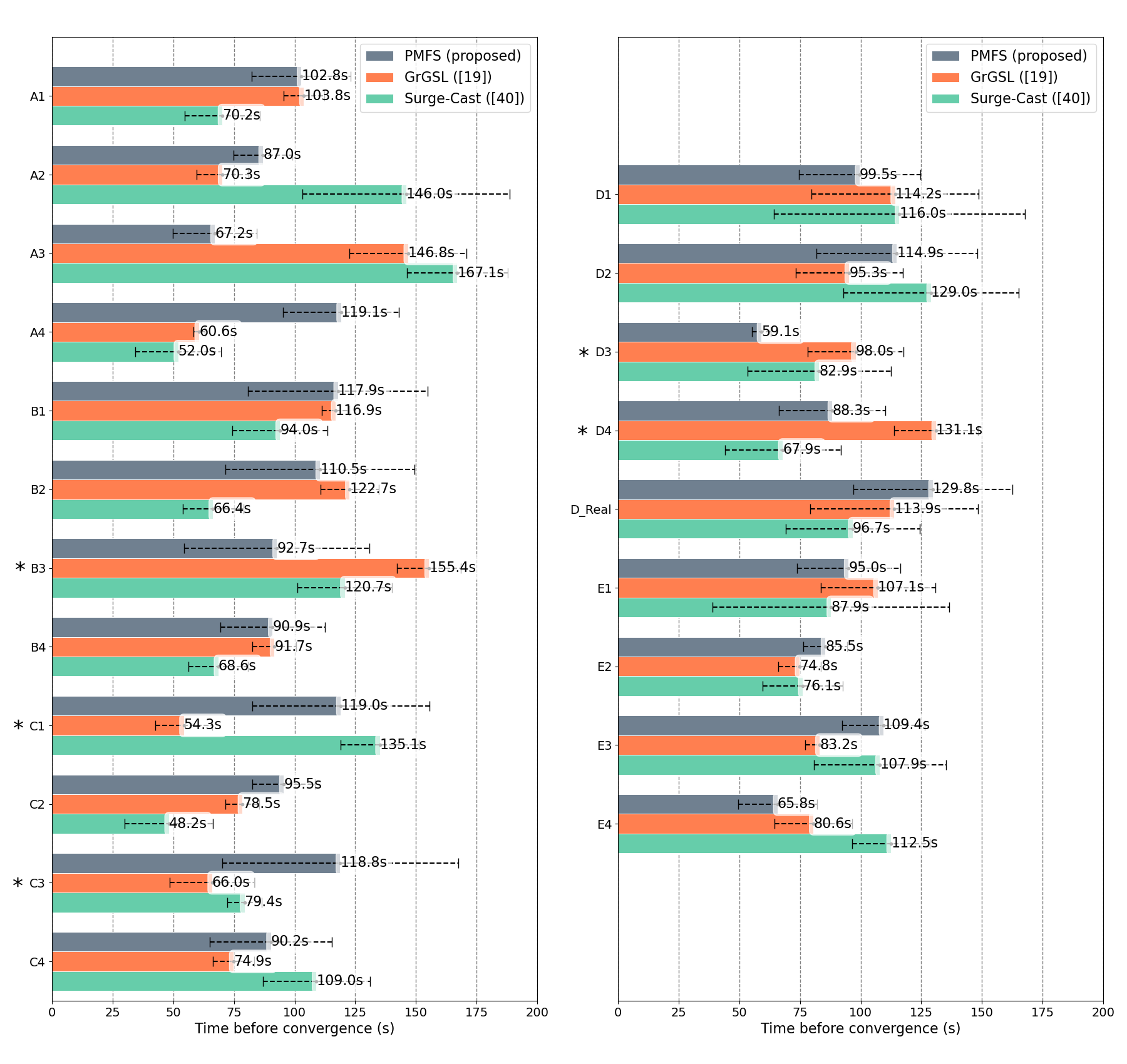}
    \caption{Time required to finish the search in each experiment. For Surge-Cast, the time it took for it to reach the smallest distance to the source, as reported in Fig.~\ref{fig:results_error}.}
    \label{fig:results_time}
\end{minipage}
\end{figure*}

\begin{figure}
    \centering
    \includegraphics[width=\linewidth]{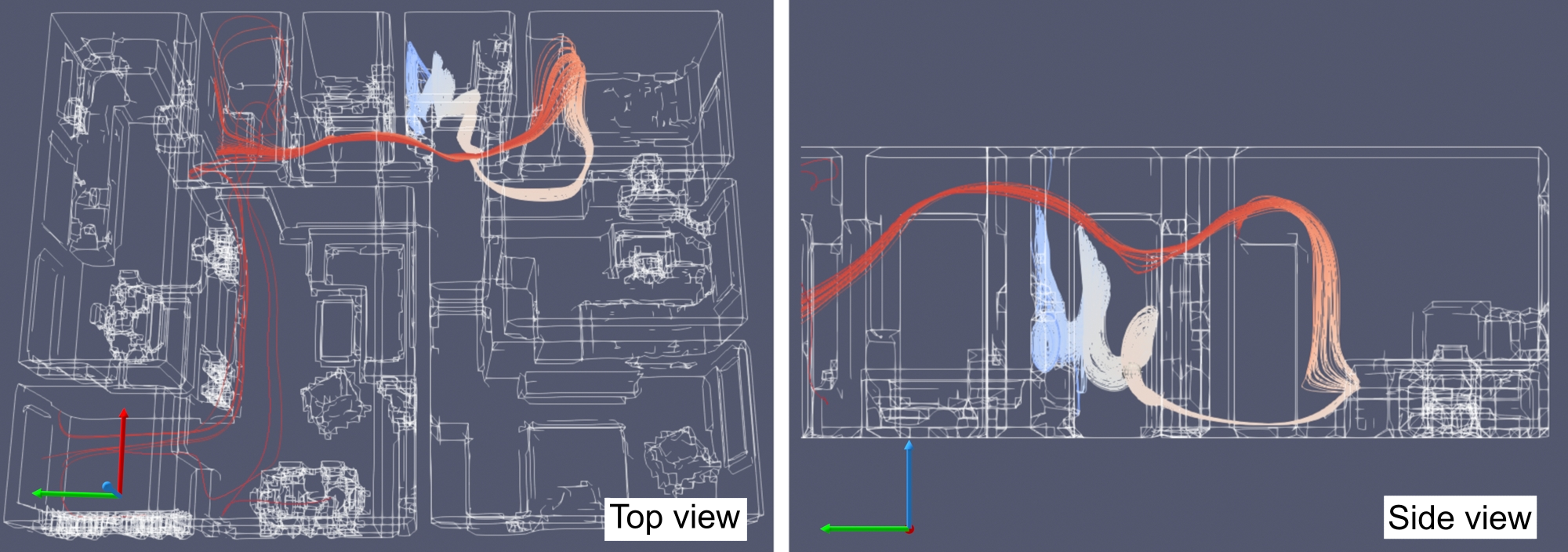}
    \caption{Stream trace of the airflow near the source position in scenario B2. The strong three-dimensionality of the dispersion process poses a challenge for the proposed method, since the dispersion model used by PMFS only considers two dimensions.}
    \label{fig:3dAirflow}
\end{figure}

\subsection{Setup}
\label{sec:experiments}
The experimental validation of the proposed algorithm was carried out by performing both simulation and real-world testing. The benefit of using simulation is that it allows for easily repeatable experimentation under a wide range of scenarios and conditions, while the real-world experiments serve to verify that the results obtained in the simulations still hold under real, uncontrolled conditions. All configuration files for these experiments (both simulated and real), which include the input parameters of the algorithm, can be found on an online repository~\footnote{\url{https://github.com/MAPIRlab/Gas-Source-Localization}}. In the case of the simulated experiments, this includes not only the configuration of the robot and algorithm, but also all the data about the airflow, inlets and outlets, source release rate, \textit{etc}.

For the simulations, we rely on GADEN~\cite{monroyGADEN3DGas2017}, a 3D gas dispersion simulator that employs CFD-based airflow. We tested the algorithm's performance under 20 different scenarios, taking place in five distinct environments (Fig.~\ref{fig:sim_experiments}). Two of these scenarios correspond to the real-world experiments. The environments are as follows:

\begin{itemize}
    \item Environment A is a simplified version of a generic indoor location, featuring multiple rooms and walls, but no limited-height obstacles to fully explore the three-dimensionality of the gas dispersion process.
    \item Environments B and C are models of real houses, and the specific scenarios considered have been selected from the VGR dataset~\cite{VGRDataset}. These scenarios include significantly more complex geometry that prevents the airflow from being consistent in the vertical axis.
    \item Environments D and E are the scenarios for the two real world experiments. Environment D is a house, with similar characteristics to B and C. Environment E is a research laboratory. Both of these environments have also been used in simulation, to compare the results obtained there with those of the real experiments.
\end{itemize}

We label experiments that take place in the same environment with a number -- \textit{e.g.} experiments A1 and A2 both take place in environment A, but with a different configuration: the source is placed at a different location and/or the airflow inlets and outlets have changed. 

We compare the currently presented algorithm (labeled in the results as \textit{PMFS}) to the algorithm presented in~\cite{Ojeda2021RAL}, (labeled \textit{GrGSL}) and Surge-Cast plume tracking~ \cite{lochmatterTrackingOdorPlumes2009}. Each of the experiments comprised 30 runs for each of the algorithms, which are considered to end once the variance of the probability distribution of the source location falls below a fixed threshold --in this case, $1m^2$. In the case of Surge-Cast, which does not have a probability distribution for the source, the algorithm is allowed to run uninterrupted for 300s.

The real-world experiments correspond to simulations D1 and E1, respectively. The gas source was an ultrasonic vibration humidifier loaded with a 96\% ethanol solution. We used a Giraff robot equipped with a photoionization detector (PID) and an ultrasonic anemometer for GSL, and with a 2D laser scanner and RGB-D camera for navigation. Because these experiments are vastly more time-consuming to carry out, only 10 runs of each algorithm were recorded as a way to validate the more extensive simulation results.

\subsection{Results}
\label{sec:experiments-results}
Results are shown in Fig.~\ref{fig:results_error}. For PMFS and GrGSL, the "error" value displayed in Fig.~\ref{fig:results_error} is the distance in meters between the ground-truth source location and the source position declared by the algorithm after reaching the convergence criterion. For Surge-Cast, since it is a purely navigational algorithm that does not have a mechanism for source declaration, the value displayed is the minimum distance between the robot position and the source achieved during the search. While this is certainly a different metric, we show it alongside the error in source declaration to provide a reference value that should give the reader an indication of the complexity of the setup --\textit{i.e.} experiments where Surge-Cast performs well can be assumed to have clear, uninterrupted gas plumes that extend to the source position.

\begin{figure}[t]
    \centering
    \includegraphics[width=\linewidth]{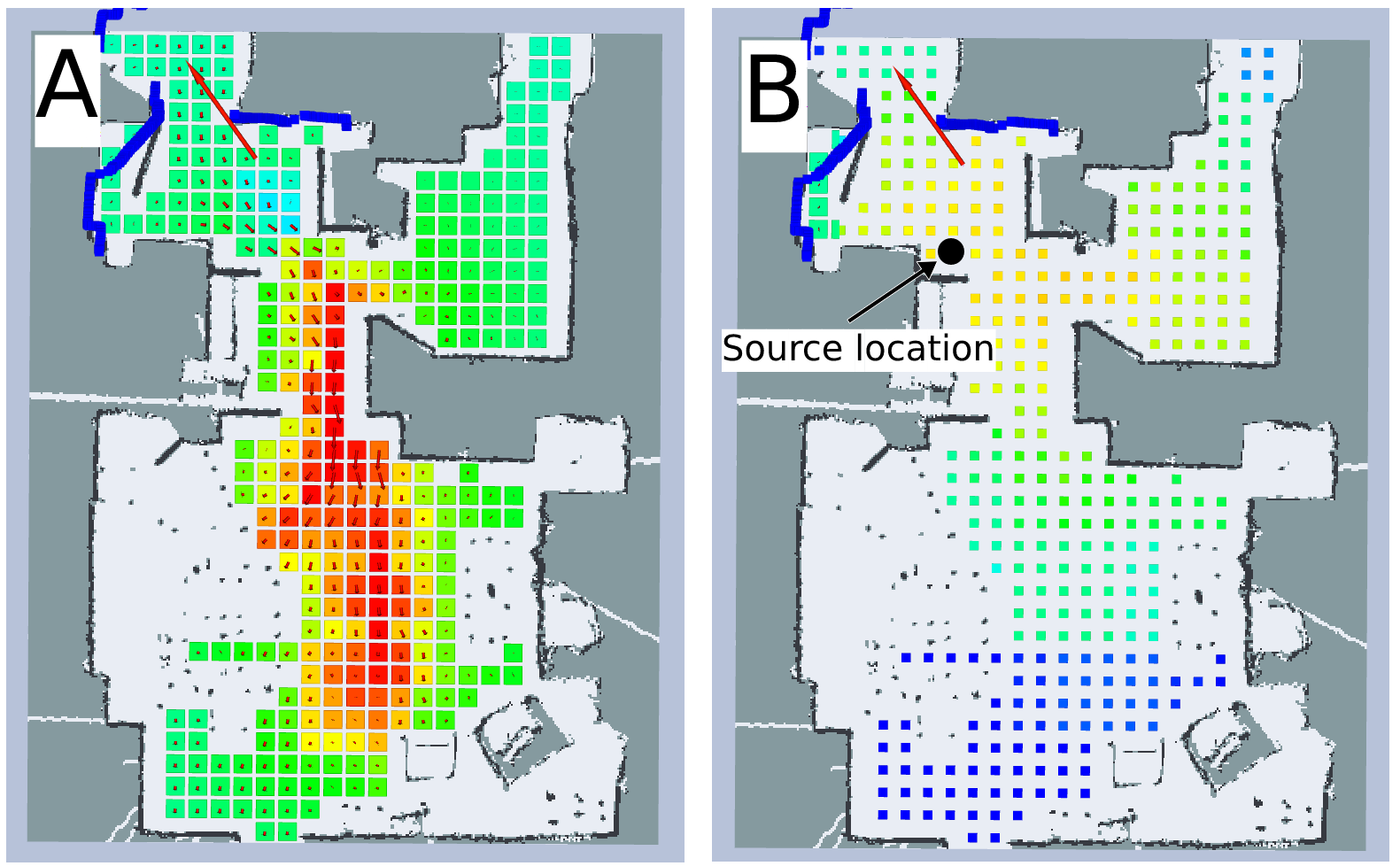}
    \caption{Snapshot of the calculated probabilities during the search in the real-world environment. (A) Hit probability map, with the arrows inside the cells showing the estimated wind direction. (B) Source location probability distribution. The red arrow is the pose of the robot, and the blue lines indicate the measurements of the laser scanner.}
    \label{fig:real_screenshot}
\end{figure}

Figure~\ref{fig:results_time} also shows the total time in seconds required for the algorithm to find the gas source. Again, in the case of Surge-Cast the lack of a source declaration mechanism means that we cannot report the same metric as with the other two algorithms, instead showing the amount of time it took to reach the closest position to the source (the one reported in Figure~\ref{fig:results_error}). Similarly to the error figure, this cannot be interpreted as a direct comparison, but rather as a baseline reference value.

It can be observed that the proposed method outperforms \textit{GrGSL} in the majority of the more complex environments, while obtaining comparable results in the simpler cases. Experiment B2 proves to be particularly challenging, with both algorithms producing estimations that are, on average, more than 2m away from the actual source position. This shows one of the main limitations of our proposed method, which is that it very strongly relies on obtaining an accurate estimation of the direction of the airflow, and is unable to do so when the three-dimensionality of the environment plays an important role (Fig.~\ref{fig:3dAirflow} shows a stream trace of the airflow around the point where the source is placed). In other cases (see scenarios B1, C1 and C2), the new method is able to produce much better estimations than the previous algorithm, because \textit{GrGSL} tends to produce an estimation that is as far upwind as possible inside the gas plume, while the reasoning of the new method (comparing the plume that would be produced if the source was in a specific location to the currently mapped plume) allows it to produce estimations that are not aligned with the airflow currents.

Regarding the time required for convergence, it can be observed that there is neither PMFS pr GrGSL manage a clear improvement over the other. There is high variance on which method achieves the best time for any given experiment, and both of them tend to be comparable to the amount of time required for a reactive navigation algorithm like Surge-Cast. Further testing with multiple movement strategies (for example, the concept discussed in Section~\ref{sec:movement-information}, where the cost of each movement is taken into consideration alongside its information value) would be required to draw more relevant conclusions about the search time.

The results obtained in the real-world experiments are coherent with what might be expected given the simulations. It can be observed that both methods obtain comparable results, both in terms of the error in the final estimation, and in the amount of time required to produce it. While the error recorded from PMFS is on average lower in both scenarios, the variance of the results and the lower repetition count of the real-world experiment compared to the simulations, make this relatively small difference not very meaningful. Figure~\ref{fig:real_screenshot} shows a snapshot of the probabilities calculated by PMFS (of each $H_i$, and of $S$) during the search in environment D. It can be observed that the area that receives the highest probability of containing the source is at the start of the plume (following it upwind), but that, given that many cells remain unobserved (with $P(H_i | Z) \approx P(H_i)$ and $\alpha_i \approx 0$) there still exists significant uncertainty in $P(S | H)$.

\section{Conclusions and Future Work}
The gas source localization algorithm presented in this work revolves around the idea of using a forward gas dispersion model to estimate the gas plume that would be produced if the source was in a certain position, and comparing that prediction with the data that has been obtained so far.

The specifics of the model used and the calculations done with those results, as presented here, should not be considered a finalized solution, but merely a first attempt at making a feasible implementation of the concept. Indeed, many of the most important limitations of the method (mainly those related to the 3D phenomena) could be addressed by using more robust, more accurate methods for the airflow estimation and the gas dispersion, although a compromise between accuracy and computational complexity will always be required to perform on-line simulations during the search.

Another limitation that should be addressed by future work is the need to know the map of the environment in advance. Developing methods for estimating the airflow and simulating the gas dispersion in partially observed environments with uncertainty would be of great importance in making the concepts presented here fit for real applications.


\small
\bibliographystyle{IEEEtran}
\bibliography{IEEEabrv,main}
\end{document}